# A Novel Dual Quaternion based Dynamic Motion Primitives for Acrobatic flight

Renshan Zhang[1], Yongyang Hu[1], Kuang Zhao[1], Su Cao[2*]
[1]Nanjing Telecommunication Technology Research Institute, Nanjing, China
[2]Institute of Unmanned Systems, National University of Defense Technology, Changsha, China
zhangrenshan_tx@126.com, caosu_93@foxmail.com

*Abstract*—The realization of motion description is a challenging work for fixed-wing Unmanned Aerial Vehicle (UAV) acrobatic flight, due to the inherent coupling problem in translational-rotational motion. This paper aims to develop a novel maneuver description method through the idea of imitation learning, and there are two main contributions of our work: 1) A dual quaternion based dynamic motion primitives (DQ-DMP) is proposed and the state equations of the position and attitude can be combined without loss of accuracy. 2) An online hardware-in-the-loop (HITL) training system is established. Based on the DQ-DMP method, the geometric features of the demonstrated maneuver can be obtained in real-time, and the stability of the DQ-DMP is theoretically proved. The simulation results illustrate the superiority of the proposed method compared to the traditional position/attitude decoupling method.

## I. INTRODUCTION

The acrobatic of UAV is usually accompanied by fast change in position and attitude, which is a relatively complex movement process[1]. Traditional methods are difficult to describe this kind of action. Imitation learning provides a way of thinking, typically the dynamic motion primitive (DMP) can represent complex motion, and guarantee the stability and continuity of the trajectory, which is a scheme worth exploring[2].

Motion primitives theory uses the sequencing of biological systems and the ability to adapt to motion units to explain the execution of complex motions. DMPs originated from the motion control of biological systems, and can be regarded as a rigorous mathematical formula for stable non-linear dynamic systems of motion primitives[3]. DMPs are a trajectory planning method proposed by Stefan Schaal in [4] and updated by Auke Ijspeert in [5]. For another Pastor et al. [6] first extended DMP to rotational motion and proposed quaternion DMP, Ude et al. [7] extended the rotation matrix quaternion DMP, Saveriano et al. [8] used the Lyapunov method to prove the stability of their system. These works extend DMP from $\mathbb{R}$ space to $SO(3)$ space.

On the application, DMPs are widely used in robotic arms, humanoid robots, medical enhancement robots, teleoperation robots, as well as UAV and other mobile robots. In the field of drones, Perk and Slotine [9] use DMP to define the flight path and obstacle avoidance of UAV, where the trajectory is generated by the movement of the joystick that controls the UAV. Later, Fang et al. [10] extended the method to encode the drone data demonstrated by the user, extract and encode the linear part of the flight trajectory, and combine them into flight control actions. In addition Tomi´c et al. [11] formulated the motion of the drone as an optimal control problem, the output of the optimal control solver is encoded using DMP, so that they could apply modifications to the UAV's flight trajectory in real time. Lee et al. [12] also incorporated DMP into the control scheme to modify the flight trajectory and avoid obstacles in flight. DMP has played an important role in avoiding real-time obstacles.

These works apply DMP to the field of UAV and have made a lot of progress, but they have not considered the coupling of position and attitude. Traditional methods usually decouple position and attitude. This method undermines the integrity of the problem, showing disadvantages and limitations that cannot be balanced. The dual quaternion uses only eight real numbers to describe the motion of a general rigid body, which can concurrently express rotation and translation as well as the coupling relationship between each other, which can realize the integration of pose in the true sense[13]. Based on quaternion DMP, this paper proposes DQ-DMP, expanding DMP to $SE(3)$ space. Finally, we apply it to describe the complex motion of fixed-wing UAV, and train DQ-DMP through expert teaching data, the simulation results show the feasibility of this method.

## II. FORMULATION OF DMPs

In this section, we introduce several theoretical models of DMP. First, we introduce the classic DMP and the quaternion DMP, and then briefly introduce the dual quaternion to provide a basis for the next section of the dual quaternion DMP.

### A. Classical DMP

For a typical second-order point attractor system, such as mass spring damping system, which can be expressed using the following equation $\ddot{y} = \alpha_z \left[ \beta_z (g - y) - \dot{y} \right]$, where $g$ is the end point of the system, $y$ is the state of the system, $\alpha$ and $\beta$ is the gain, the control goal is to make the system reach the specified end point. In order to make the

system move according to the trajectory we expect, the phase variable and forcing term are introduced[5]

$$\tau \dot{y} = z$$
$$\tau \dot{z} = \alpha_z \left[ \beta_z (g - y) - z \right] + f(x) \quad (1)$$

where the phase variable is the exponentially decayed clock signal from 1 to 0, obtained by the so-called regular system integration $\tau \dot{x} = -\alpha_x x$, time scale factor $\tau$ can change the duration of motion, $f(x)$ is defined as the weighted average of $N$ Gaussian kernel functions, its function is to change the acceleration of the system at different moments, and drive the system along a freewill smooth trajectory from the initial position $y_0$ to the end point $g$, where forcing term

$$f(x) = \frac{\sum_{i=1}^{N} \lambda_i \psi_i(x)}{\sum_{i=1}^{N} \psi_i(x)} x \quad (2)$$
$$\psi_i(x) = \exp\left(-h_i (x - c_i)^2\right)$$

where $c_i$ is the center of Gaussian functions distributed along the motion phase, and $h_i$ is their width. For a given $N$, set $\tau$ equal to the total duration of the desired movement. Generally speaking, we can define parameters

$$c_i = \exp\left(-\alpha_x \frac{i-1}{N-1}\right), h_i = \frac{1}{(c_{i+1} - c_i)^2}, h_N = h_{N-1}. \quad (3)$$

For each degree of freedom, the weight $\lambda_i$ should be adjusted according to the measured data in order to achieve the desired behavior. For a given trajectory $y_d(t)$ and its derivative $\dot{y}_d(t)$ and $\ddot{y}_d(t)$, $t=0, \Delta t, \ldots, T$, the required values $f_d(t)$ can be obtained as follows

$$f_d(t) = \tau^2 \ddot{y}_d(t) - \alpha_z \left[ \beta_z (g - y_d(t)) - \tau \dot{y}_d(t) \right]. \quad (4)$$

We use $f(x)$ formed by basis function to approach the target driving term, the driving term and weight can be approximated as a linear relationship, that is $f(x) = A \times \lambda \approx f_d(t)$.

$$A = \begin{bmatrix} \frac{\psi_1(x_1)}{\sum_{i=1}^{N} \psi_i(x_1)} x_1 & \cdots & \frac{\psi_N(x_1)}{\sum_{i=1}^{N} \psi_i(x_1)} x_1 \\ \vdots & \ddots & \vdots \\ \frac{\psi_1(x_T)}{\sum_{i=1}^{N} \psi_i(x_T)} x_T & \cdots & \frac{\psi_N(x_T)}{\sum_{i=1}^{N} \psi_i(x_T)} x_T \end{bmatrix} \quad (5)$$

It can be obtained that $\lambda = A^\dagger f_d(t)$, $A^\dagger$ is the pseudo-inverse of $A$, and the parameter $\lambda_i$ can be calculated by the weighted least square method.

It should be noted that the above system defaults choose the inertial coordinate system.

For controlling a robot system with multiple degrees of freedom, we use a different system of equations (1) to express the motion of each degree of freedom, but use a common phase to synchronize them. It can be seen that the dynamic motion primitive is driven by the attractor dynamics differential equation, and represented by a combination of the nonlinear force term and the attractor force term. Nonlinear force can represent complex motion, and attractor force represents the target state. The nonlinear force weakens with time, and finally the attractor force dominates, so the dynamic motion primitive can smoothly converge to the target state. DMPs ensure the smoothness and continuity of the trajectory, and can express nonlinear motion without losing stability.

### B. Quaternion DMP

In Cartesian space, attitude can be expressed as a unit quaternion $q = [\eta, \varepsilon] \in SO(3)$, where $SO(3)$ is the unit ball of three-dimensional space, and $\eta$ is scalar part, $\varepsilon$ is the vector part of the quaternion, $\eta^2 + \|\varepsilon\|^2 = 1$. Compared with the rotation matrix, the unit quaternion has fewer parameters, and there is no singularity in contrast to Euler angles, so it is frequently used to describe rotation operations in engineering.

The quaternion DMP of the inertial coordinate system is expressed as follows[7]

$$\tau \dot{q} = \frac{1}{2} \tilde{\omega} \otimes q$$
$$\tau \dot{\omega} = K^o \left[ e_o (q_d \otimes q^*) - d_0^q(x) + f^q(x) \right] - D^o \omega \quad (6)$$

where $\omega \in \mathbb{R}^3$ is the angular velocity, $\dot{\omega} \in \mathbb{R}^3$ is the angular acceleration, $d_0^q(x) = (q_d - q_0) x$ prevent jumping at the beginning, $f^q(x)$ is obtained by (2). $\tilde{\omega} = [0, \omega]$ is a quaternion, the scalar part is zero, and the vector part is the angular velocity. The error between the two quaternions $e_o(\cdot, \cdot)$ is a nonlinear function, it has multiple definitions, in this work we choose the orientation error between $q_1$ and $q_2$ as $e_o = \text{vec}(q_1 \otimes q_2^*)$ [8], where $\text{vec}(q)$ back to the vector part of $q$.

Define the differential of a quaternion

$$q(t + \Delta t) = \exp\left(\frac{\Delta t}{2} \omega(t)\right) \otimes q(t) \quad (7)$$

where $\Delta t$ is the sampling time, so $q$ can be calculated by the above formula, where the exponential mapping of the quaternion $\exp(\cdot): \mathbb{R}^3 \mapsto SO(3)$, a rotation vector is exponentially mapped to a unit quaternion

$$\exp(r) = \begin{cases} \left[\cos(\|r\|), \sin(\|r\|) \frac{r}{\|r\|}\right], & r \neq 0 \\ [1, 0, 0, 0]^T, & \text{otherwise} \end{cases} \quad (8)$$

## C. Introduction to dual quaternions

The dual number was invented by Clifford in 1873 and further expanded by Study in 1891. The dual number is defined as

$$\hat{z} = a + \epsilon b \text{ with } \epsilon^2 = 0, \text{ but } \epsilon \neq 0 \quad (9)$$

where $a$ and $b$ are real numbers, called the real part and the dual part respectively, and $\epsilon$ is the nilpotent term.

A dual quaternion $\hat{q} = q_o + \epsilon q_p \in SE(3)$ is a dual number whose elements are quaternion, where $q_o$ is the real part, representing rotation, $q_p$ is an even part, representing translation, which are all quaternion and $\epsilon$ is nilpotent term. This method of representation can unify the translation and rotation in one space, the general rigid body motion described by the rotation $q$ followed by the translation $p^b$ can be described as $\hat{q} = q + \frac{\epsilon}{2} q \otimes p^b$, where $p^b$ represents a position in the body coordinate system[13].

The error of the dual quaternion can be expressed as $\hat{q}_e = q_{oe} + \epsilon q_{pe} = q_{oe} + \frac{\epsilon}{2} q_{oe} \otimes p_e^b$, and we can get $p_e^b = 2 q_{oe}^* \otimes q_{pe}$.

The kinematic equation of dual quaternion is

$$\begin{aligned} \dot{\hat{q}} &= \frac{1}{2} \tilde{\xi}^s \otimes \hat{q} \\ \dot{\hat{q}} &= \frac{1}{2} \hat{q} \otimes \tilde{\xi}^b \end{aligned} \quad (10)$$

where $p^b$ is the quaternion expanded by adding zero elements at the position, $\tilde{\xi}$ is the dual quaternion expanded by spinor $\xi \in se(3)$.

$$\begin{aligned} \xi^s &= \omega^s + \epsilon(\dot{p}^s + \omega^s \times p^s) \\ \xi^b &= \omega^b + \epsilon(\dot{p}^b + p^b \times \omega^b) \end{aligned} \quad (11)$$

They are the spinor in the inertial coordinate system and the spinor in the body coordinate system, $p^s$ and $p^b$ are the projections of the translation motion contained in the dual quaternion.

## III. DUAL QUATERNION BASED DMPs

### A. Dual quaternion DMP

The dual quaternion DMP in the body coordinate system is expressed as follows

$$\begin{aligned} \tau \dot{\hat{q}} &= \frac{1}{2} \hat{q} \otimes \tilde{\xi}^b \\ \tau \dot{\xi}^b &= K^{\hat{q}} \left[ e_o \left( \hat{q}^* \otimes \hat{q}_d \right) - d_0^{\hat{q}}(x) + f^{\hat{q}}(x) \right] - D^{\hat{q}} \xi^b \end{aligned} \quad (12)$$

where $\tau$ is the time scale, $d_0^{\hat{q}}(x) = (\hat{q}_d - \hat{q}_0) x$ prevent jumping at the beginning, $f^{\hat{q}}(x)$ is defined by (2), $K^{\hat{q}}$, $D^{\hat{q}}$ is linear stiffness gain and damping gain , including attitude part and position part.

For a given trajectory $\hat{q}_d(t)$, the forcing term is

$$f^{\hat{q}}(t) = \left( K^{\hat{q}} \right)^{-1} \left( \tau^2 \dot{\xi}^b + \tau D^{\hat{q}} \xi^b \right) - e_o \left( \hat{q}^* \otimes \hat{q}_d \right) + d_0^{\hat{q}}(x) \quad (13)$$

$f^{\hat{q}}(t)$ can be used to encode any sampled directional trajectory $\left[ \hat{q}_j, \xi_j, \dot{\xi}_j, t \right]_{j=0}^T$, and weight $\lambda_i$ can be obtained by solving the following linear equations

$$\frac{\sum_{i=1}^N \lambda_i \psi_i(x)}{\sum_{i=1}^N \psi_i(x)} x = \left( K^{\hat{q}} \right)^{-1} \left( \tau^2 \dot{\xi}^b + \tau D^{\hat{q}} \xi^b \right) - e_o \left( \hat{q}^* \otimes \hat{q}_d \right) + d_0^{\hat{q}}(x). \quad (14)$$

In the calculation process, it is necessary to define the orientation error between the dual quaternions $\hat{q}$ and $\hat{q}_d$, this paper adopts the following logarithmic mapping $\ln(\cdot): SE(3) \mapsto se(3)$

$$e_o \left( \hat{q}^* \otimes \hat{q}_d \right) = \ln(\hat{q}_e) = \left[ \text{vec}(q_{oe}), \text{vec}(p_e) \right]. \quad (15)$$

Using kinematic equations, we can define the dual differential quaternion as follows

$$\hat{q}(t + \Delta t) = \hat{q}(t) \otimes \exp\left( \frac{\Delta t}{2} \xi^b(t) \right). \quad (16)$$

And set $\xi = [r, v]$, we define the exponential mapping of the spinor as $\exp(\cdot): se(3) \mapsto SE(3)$

$$\exp(\xi) = \begin{cases} \left[ \cos(\|r\|), \sin(\|r\|) \frac{r^T}{\|r\|} \right]^T + \epsilon [0, v]^T, r \neq 0 \\ [1, 0, 0, 0]^T + \epsilon q_d, \quad \text{otherwise.} \end{cases} \quad (17)$$

### B. Stability analysis

When $t \to \infty$, $d_0^{\hat{q}}(x), f^{\hat{q}}(x)$ gradually vanish with the $x$ decay from 1 to 0, the system degenerates to $\tau \dot{\xi}^b = K^{\hat{q}} e_o \left( \hat{q}^* \otimes \hat{q}_d \right) - D^{\hat{q}} \xi^b$.

**Theorem 1.** *For the above dual quaternion DMP system, $x$ is a clock signal with exponential decay from 1 to 0 with time. The error is defined as $\left[ \text{vec}(q_{oe}), \text{vec}(p_e) \right]$, the system globally asymptotically converges to $\hat{q} = \hat{q}_d$ with $\xi^b = 0$. If the state quantity is selected $x = [\hat{q}, \xi^b]$, it is $x$ close to $[\hat{q}_d, 0]$*

*Proof:* For a unified description, the following derivation selects the body coordinate system, without loss of generality, we set $\tau = 1$ and define $\xi^b = [\omega, v]$.

$$\begin{aligned} \dot{\xi}^b &= K^{\hat{q}} e_o \left( \hat{q}^* \otimes \hat{q}_d \right) - D^{\hat{q}} \xi^b \\ &= \begin{bmatrix} K^o & 0 \\ 0 & K^p \end{bmatrix} \begin{bmatrix} \text{vec}(q_e) \\ \text{vec}(p_e) \end{bmatrix} - \begin{bmatrix} D^o & 0 \\ 0 & D^p \end{bmatrix} \begin{bmatrix} \omega \\ v \end{bmatrix}. \end{aligned} \quad (18)$$

We can get the following formula
$$\dot{\omega} = K^o vec(q_o^* \otimes q_{od}) - D^o \omega$$
$$\dot{v} = K^p vec(p_d - p) - D^p v \quad (19)$$

If set $\hat{q}_d = q_{od} + \epsilon q_{pd} = [\eta_{od}, \varepsilon_{od}] + \dfrac{\epsilon}{2} q_{od} \otimes p$, we can use the following Lyapunov candidate

$$V(x) = (\eta_{od} - \eta_o)^2 + \|\varepsilon_{od} - \varepsilon_o\|^2 + \dfrac{1}{2}\omega^T (K^o)^{-1} \omega \quad (20)$$
$$+ \dfrac{1}{2}\|p_d - p\|^2 + \dfrac{1}{2} v^T (K^q)^{-1} v$$

Considering that the rotations is a fast item and is not affected by the translation, $V(x)$ will be split as follows
$V(x) = V_1(x) + V_2(x)$.

where $V_1(x) = (\eta_{od} - \eta_o)^2 + \|\varepsilon_{od} - \varepsilon_o\|^2 + \dfrac{1}{2}\omega^T (K^o)^{-1} \omega$ and
$V_2(x) = \dfrac{1}{2}\|p_d - p\|^2 + \dfrac{1}{2} v^T (K^q)^{-1} v$.

For $V_1(x)$, take the derivative of both sides
$$\dot{V}_1(x) = -2(\eta_{od} - \eta_o)\dot{\eta}_o - 2(\varepsilon_{od} - \varepsilon_o)^T \dot{\varepsilon}_o + \omega^T (K^o)^{-1} \dot{\omega} \quad (21)$$

By the formula (19) and (28)
$$\dot{V}_1(x) = (\eta_{od} - \eta_o)\omega^T \varepsilon - (\varepsilon_{od} - \varepsilon_o)^T (\eta_o \omega + \varepsilon_o^\times \omega)$$
$$+ \omega^T (K^o)^{-1} (K^o vec(q_o^* \otimes q_{od}) - D^o \omega)$$
$$= -\omega^T (K^o)^{-1} D^o \omega + \omega^T vec(q_o^* \otimes q_{od}) \quad (22)$$
$$+ \omega^T (\eta_{od}\varepsilon - \eta_o\varepsilon_{od} + \varepsilon_o^\times \varepsilon_{od})$$
$$= -\omega^T (K^o)^{-1} D^o \omega + \omega^T (vec(q_o^* \otimes q_{od}) - vec(q_o^* \otimes q_{od}))$$
$$= -\omega^T (K^o)^{-1} D^o \omega$$

Design $K^o, D^o$ to be positive definite can ensure that $\dot{V}_1(x)$ is negative definite, so $q_o$ asymptotically converges to $q_{od}$, with $\omega = 0$.

$$\dot{V}_2(x) = -(p_d - p)^T \dot{p} + v^T (K^q)^{-1} \dot{v}$$
$$= -(p_d - p)^T \dot{p} + v^T (K^q)^{-1} (K^p vec(p_d - p) - D^p v)$$
$$= v^T vec(p_d - p) - v^T (K^q)^{-1} D^p v - (p_d - p)^T (v - p \times \omega) \quad (23)$$
$$= (p_d - p)^T (p \times \omega) - v^T (K^q)^{-1} D^p v$$
$$= p_d^T (p \times \omega) - v^T (K^q)^{-1} D^p v$$
$$= \omega^T (p \times p_d) - v^T (K^q)^{-1} D^p v$$

Under the premise of $q_o$ asymptotic convergence, $\omega$ converges to 0, $\omega^T (p \times p_d)$ will be an item as small as possible, so $\dot{V}_2(x)$ can be inferred negative definite, $p$ asymptotically converges to $p_d$, with $v = 0$.

To sum up, the system globally asymptotically converges to $\hat{q}_d$ with $\xi = 0$.

## IV. SIMULATION DESIGN

The simulation system is built by the open source flight control px4 and the flight simulation software X-Plane. The joystick sends instructions to the flight control, after processing, the control signal is sent to X-Plane to control the aircraft in X-Plane. This is a typical hardware-in-the-loop simulation system.

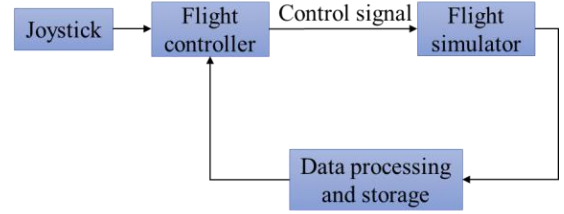

Figure 1. Data acquisition system

For different maneuvers, we collect flight data through expert teaching, and the sampling frequency is set to 100Hz. This data is used to train the DMP, the entire training process is shown in the following figure.

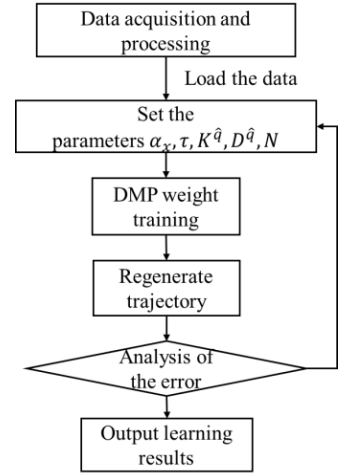

Figure 2. DMP learning process

The Gaussian kernel function is selected as
$$\psi_i(x) = \dfrac{\exp\left(-\dfrac{1}{2}(x - c_i)^2 h_i^{-1}\right)}{\sqrt{2\pi h_i}} \quad (24)$$

Set total duration as $T$ and sampling time as $\Delta t$
$$c_i = \exp\left(-\alpha_x T \dfrac{i}{N+1}\right), i = 1, \ldots, N$$
$$h_i = \left[c_i - \exp\left(-\alpha_x \left(1 + \dfrac{(T - \Delta t)(N - i)}{\Delta t(N - 1)} + \sqrt{10 \dfrac{T}{\Delta t}}\right)\Delta t\right)\right]^2 \quad (25)$$

In this experiment, the somersaults motion is taken for a simulation test, which lasts 18.9 seconds, the final parameters are selected as follows

Pose DMP(Consider both position and quaternion): $\alpha_x = 0.1, N^p = 30, K^p = 10, D^p = 10\sqrt{K^p}, N^o = 50, K^o = 1$ and $D^o = 10\sqrt{K^o}$.

Dual quaternion DMP: $\alpha_x = 0.05, N^{\hat{q}} = 30, K^p = 1$, and $D^p = 10\sqrt{K^p}, K^o = 1, D^o = 10\sqrt{K^o}$.

## V. RESULTS AND DISCUSSION

We show the results of the training in Figure 3 to 4. It should be noted that position of traditional pose DMP is only applicable to the inertial coordinate system. However, the reference frame of angular velocity is the body coordinate, because it is generally selected by default in the data collection of the flight controller.

For the quaternion DMP in the Sec. II, the formula (6) and (7) needs to be revised as follows in the body coordinate system

$$\tau \dot{q} = \frac{1}{2} q \otimes \tilde{\omega}$$
$$\tau \dot{\omega} = K^o \left[ e_o \left( q^* \otimes q_d \right) - d_0^q(x) + f^q(x) \right] - D^o \omega \quad (26)$$
$$q(t + \Delta t) = q(t) \otimes \exp\left( \frac{\Delta t}{2} \omega(t) \right)$$

For dual quaternion operations, a unified coordinate system is required. Here, the body coordinate system is selected for operation. In addition, the position and speed of the inertial coordinate system can be obtained through coordinate system conversion. In the training process and result display, different forms of scale transformation are performed on the position.

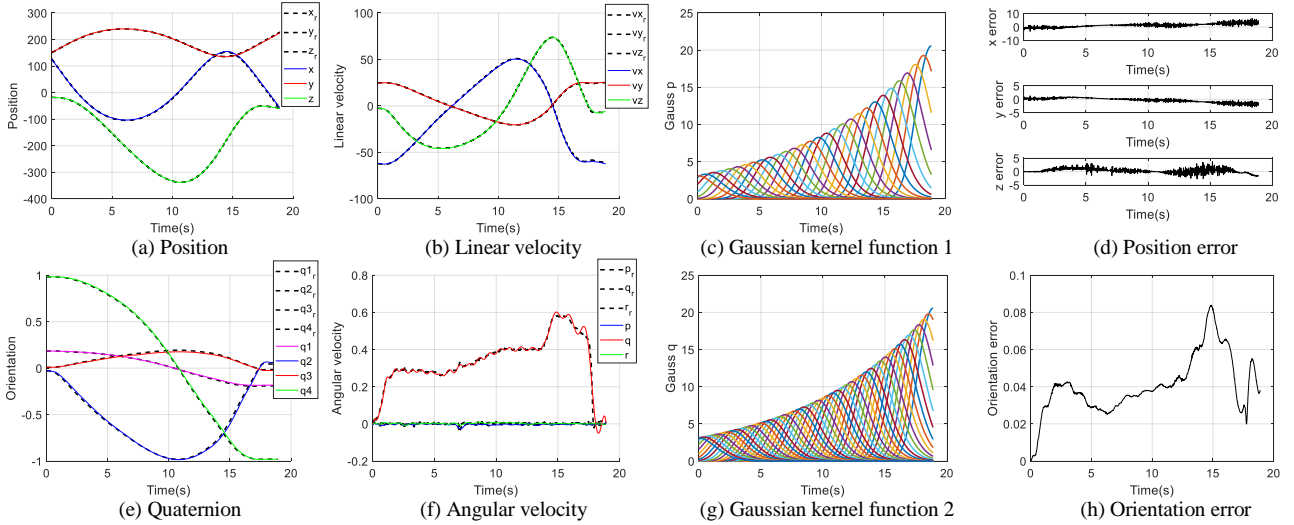

Figure 3. This figure shows the training effect of pose DMP, the result in (c) and (g) is the change of the Gaussian kernel function $\psi_i(x)$ and the error of the attitude in (h) is the norm of the quaternion, Gaussian kernel function 1 is used for training position and Gaussian kernel function 2 for attitude.

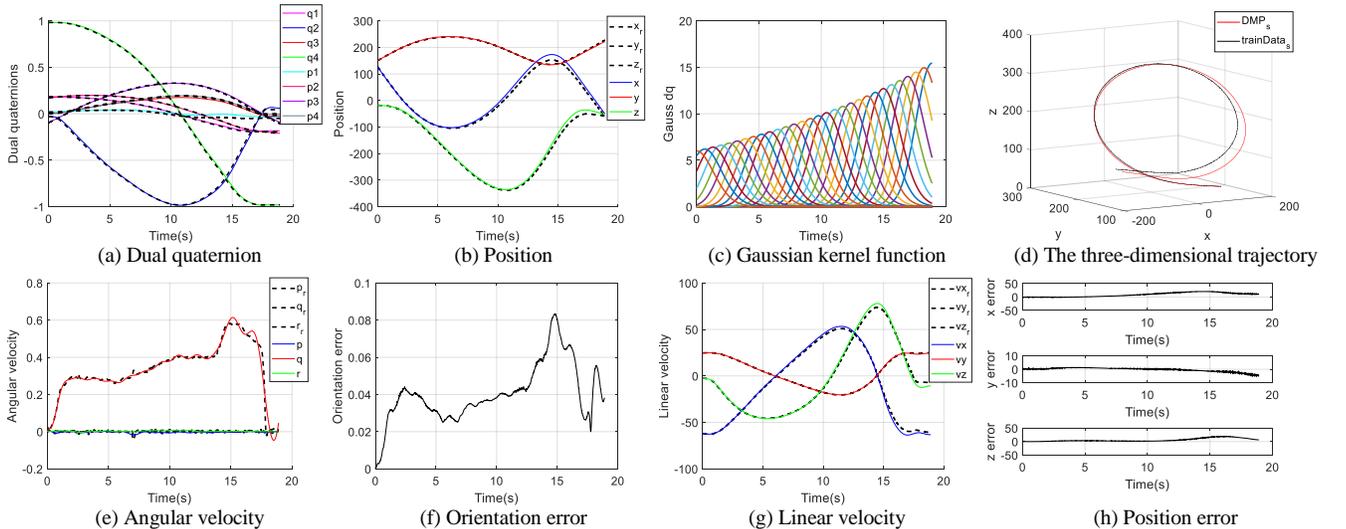

Figure 4. This figure demonstrates the training effect of Dual quaternion DMP, the definition of each quantity is the same as before.

It can be seen that the traditional pose DMP can achieve better results in their respective learning. However, such a system ignores the influence of attitude on position. During somersaults motion, the pitch angle and pitch angle velocity change fast, and the coupling of pose is serious. Obviously, in our dual quaternion DMP, when the attitude learning has a certain deviation, the position will also have a corresponding deviation, which is a performance of the UAV's motion constraint. Therefore, in the traditional pose decoupling DMP, it is more likely to train an unreachable state.

For the UAV system, it has to meet certain constraints of motion differential obviously. In this case, the dual quaternion DMP is more appropriate to describe the problem.

## VI. CONCLUSION

In this paper, through the idea of imitation learning, the motion description of the acrobatic flight of the UAV is realized. Considering the coupling problem of the position and the attitude in the acrobatic flight, the dual quaternion is introduced to describe the translational and rotational motions in a unified manner. On the basis of dual quaternion DMP, the coupling problem of position and attitude is solved. Our method was successfully applied to the acrobatic flight in a hardware-in-the-loop simulation.

In the future we will strive to express more complex maneuvers through DMP.

## APPENDIX

The basic operations of quaternions and dual quaternions are listed below.

The product of two quaternions is

$$q_1 \otimes q_2 = \left[ \eta_1\eta_2 - \varepsilon_1^T \varepsilon_2, \left( \eta_1\varepsilon_2 + \eta_2\varepsilon_1 + \varepsilon_1^\times \varepsilon_2 \right) \right] \quad (27)$$

where $\varepsilon_1^\times$ is the skew-symmetric matrix corresponding to $\varepsilon_1$. Conjugation of quaternion: $q^* = [\eta, -\varepsilon]$, $q \otimes q^* = [1, 0, 0, 0]$.

The time derivative of a quaternion can be expanded into

$$\dot{q} = \frac{1}{2} q \otimes \tilde{\omega} = \frac{1}{2}[\eta, \varepsilon] \otimes [0, \omega]$$
$$= [\dot{\eta}, \dot{\varepsilon}] = -\frac{1}{2}\varepsilon^T \omega + \frac{1}{2}\left( \eta\omega + \varepsilon^\times \omega \right). \quad (28)$$

Dual quaternions have many similar properties to ordinary quaternions. The conjugate operation of the dual quaternion selected in this paper $\hat{q}^* = q_o^* + \epsilon q_p^*$, and the addition operation of the dual quaternion is $\hat{q}_1 + \hat{q}_2 = q_{o1} + q_{o2} + \epsilon\left( q_{p1} + q_{p2} \right)$.

The multiplication operation of two dual quaternions is

$$\hat{q}_1 \otimes \hat{q}_2 = q_{o1} \otimes q_{o2} + \epsilon\left( q_{o1} \otimes q_{p2} + q_{p1} \otimes q_{o2} \right). \quad (29)$$

The time derivative of a dual quaternion can be expanded into

$$\dot{\hat{q}} = \begin{bmatrix} \dot{q}_o \\ \dot{q}_p \end{bmatrix} = \frac{1}{2}\hat{q} \otimes \xi^b = \frac{1}{2}\left( q_o + \epsilon q_p \right) \circ \left( \tilde{\omega} + \epsilon \tilde{v} \right)$$
$$= \frac{1}{2} q_o \otimes \tilde{\omega} + \frac{1}{2}\epsilon\left( q_o \otimes \tilde{v} + q_p \otimes \tilde{\omega} \right). \quad (30)$$

The rotation matrix for linear velocity from body coordinate system to inertial coordinate system is

$$\begin{bmatrix} \dot{x} \\ \dot{y} \\ \dot{z} \end{bmatrix} = \begin{bmatrix} \begin{pmatrix} q_0^2 + q_1^2 \\ -q_2^2 - q_3^2 \end{pmatrix} & 2(q_1q_2 - q_0q_3) & 2(q_1q_3 + q_0q_2) \\ 2(q_1q_2 + q_0q_3) & \begin{pmatrix} q_0^2 - q_1^2 \\ +q_2^2 - q_3^2 \end{pmatrix} & 2(q_2q_3 - q_0q_1) \\ 2(q_1q_3 - q_0q_2) & 2(q_2q_3 + q_0q_1) & \begin{pmatrix} q_0^2 - q_1^2 \\ -q_2^2 + q_3^2 \end{pmatrix} \end{bmatrix} \begin{bmatrix} u \\ v \\ w \end{bmatrix}$$

where quaternion $q = [q_0, q_1, q_2, q_3]$.